\documentclass[runningheads]{llncs}
\usepackage{graphicx}
\begin{document}
\title{A Comparison of Synthetic Oversampling Methods for \\Multi-class Text Classification
\thanks{Supported by the Russian Foundation for Basic Research (project no. 18-37-00272).}
}
\titlerunning{A Comparison of Synthetic Oversampling Methods for Multi-class Text Classification}
%
\author{Anna Glazkova\inst{1}\orcidID{0000-0001-8409-6457}}
\authorrunning{A. Glazkova}
%
\institute{Institute of Mathematics and Computer Sciences, University of Tyumen, Perekopskaya str. 15a, 625003 Tyumen, Russian Federation\\
\email{anna\_glazkova@yahoo.com}}
\maketitle              
\begin{abstract}
The authors compared oversampling methods for the problem of multi-class topic classification. The SMOTE algorithm underlies one of the most popular oversampling methods. It consists in choosing two examples of a minority class and generating a new example based on them. In the paper, the authors compared the basic SMOTE method with its two modifications (Borderline SMOTE and ADASYN) and random oversampling technique on the example of one of text classification tasks. The paper discusses the k-nearest neighbor algorithm, the support vector machine algorithm and three types of neural networks (feedforward network, long short-term memory (LSTM) and bidirectional LSTM). The authors combine these machine learning algorithms with different text representations and compared synthetic oversampling methods. In most cases, the use of oversampling techniques can significantly improve the quality of classification. The authors conclude that for this task, the quality of the KNN and SVM algorithms is more influenced by class imbalance than neural networks.


\end{abstract}
\section{Introduction}
Systematization of information is one of the main goals of natural language processing (NLP). Modern information repositories allow us to store large volumes of text documents. These documents contain a lot of information relevant to certain groups of users. Text search and classification tools need further improvement and development in order for necessary information to be accessible to users of electronic repositories.

In this paper, we focus on multi-class topic classification of text fragments containing biographical information using supervised learning methods. We use sentences as text fragments because sentences are minimum logical finished parts of the text. The problem of automatically searching for biographical information is one of the particular tasks of information retrieval. Biographical information extraction is performed not only when building a text of biographical information in search engines, but also when conducting biographical research involving working with facts relating to human life. Biographical facts relate to various aspects of life (social, political, private and etc.). At the same time, the number of text fragments relating to various topics of facts, as a rule, is not equal within one text source. In this case, any machine learning method would be biased towards the majority class.

To overcome the imbalance of classes, we use oversampling methods. In this paper, we do not aim to find the best models for multi-class classification and text representation. We want to evaluate how the oversampling methods help to improve the imbalance situation and how strongly the results of different models differ during the use and non-use of oversampling.

\section{Related Works}

\subsection{Biographical Information Retrieval}

The issues of extracting facts from a natural language text and their classification has received a growing interest in the last ten years. Scientific research in this field are characterized by different learning settings. Supervised machine learning methods are mainstream approaches in this case. For example, the extraction of biographical facts from historical texts is discussed in \cite{Adamovich,Park}. In \cite{Santos}, the authors solve the problem of classifying relations between words in a biographical text using a convolutional neural network that performs classification by ranking. The article \cite{Meerkamp} is dedicated to the architecture of a information extracting system, which combines text parsing and neural network capabilities. Most researchers use text corpora to train their models or sets of pre-selected attributes.

\subsection{Imbalanced Classification}

The problem of class imbalances is a frequent challenge in machine learning. Since the classifier in this case gives preference to the majority class, diverse studies to overcome the imbalance were conducted. Imbalance learning for different natural language processing fields has been studied by several research groups using various tasks and approaches \cite{Homma,Ah-Pine,Xu,Suh}. Unfortunately, we did not find such studies for Russian texts. In addition, most researchers compare oversampling algorithms using a single machine learning method.

\section{Methods}

\subsection{Classification Methods}

In our work, we used several popular machine learning techniques to classify texts:
\begin{itemize}
    \item k-nearest neighbor algorithm  (KNN);
    \item support-vector machine algorithm (SVM);
    \item feedforward network (FNN);
    \item long short-term memory network (LSTM);
    \item bidirectional LSTM (BLSTM).
\end{itemize}

We compared the effect of oversampling techniques over five
supervised methods. Each method uses a different approach to classification. The features of these approaches largely determine the oversampling effect on the results.

Thus, the KNN algorithm makes a decision regarding a new object based on the classes of its nearest neighbors. The lack of the nearest neighbors leads to the fact that the algorithm often assigns a new object to the majority class. 

The SVM algorithm constructs a hyperplane to classify vectors in a high-dimensional space. The optimal marginal data for the hyperplane is known as support vectors. The algorithm tries to find the optimal hyperplane that would maximize distance between the margins. The training sample vectors are used in the construction of the separating hyperplane \cite{Suh}. Accordingly, the influence of imbalance should be also significant, but probably not as decisive as in the case of the KNN algorithm.

Neural networks are a set of algorithms based on neural models representing an interconnected group of artificial neurons. It can be expected that the methods of calculating errors in the training of neural networks can reduce the impact of imbalance in comparison with previous algorithms. Despite this, research in this area shows that the effect of class imbalance on classification performance is detrimental. The minority class examples can be identified by neural network as noise, and therefore they could be wrongly discarded by the classifier \cite{Lopez,Buda}.

We consider three types of neural networks. The first one is feedforward network which is the classic architecture for neural models. The feedforward network can accept a numerical vector as input and allows using all oversampling methods for input vectors. Recurrent networks are currently showing some of the best results in the field of natural language text classification. Earlier, we compared various types of neural networks for multiclass classification of biographical text fragments \cite{article1}. The best results were achieved using LSTM- and BLSTM- networks.

\subsection{Text Representation}

We used the following ways of text representation:
\begin{itemize}
    \item Bag-of-Words;
    \item Bag-of-Words + TFIDF;
    \item Word2Vec.
\end{itemize}

The Bag-of-Words model represents a text collection as a matrix. The number of rows in the matrix is equal to the number of texts, and the number of columns is equal to the number of words in the collection (except for the list of stop words).

The Bag-of-Words + TFIDF model is similar to the previous one,  except that the intersection of the row and column contains the TFIDF value for the word in the current document.

Word2Vec \cite{Mikolov} is currently one of the most popular and effective ways to obtain word embeddings suitable for machine learning. This method is based on the frequency of words within the same context.

In our work, we used pre-trained word embeddings constructed from a snapshot of the Russian National Corpus \cite{RNC} and Russian Wikipedia \cite{Wiki} in December 2017 provided by RusVectores \cite{Kutuzov}. The snapshot contains about 600 millions words. Word vectors are obtained using Skipgram algorithm. The vector size is 300.

To create a text vector $T$ based on the words of this text $W=(w_1,w_2,...,w_n)$ and the word embedding dictionary $V=(v_{w_1},v_{w_n},...,v_{w_n})$, we applied a linear combination approach:

\begin{equation}
    \it{T} = \sum\limits_{i=1}^n\it{v_{w_i}}.
\end{equation}

Thus, we used three types of text representation for KNN, SVM and FNN models: Bag-of-Words, Bag-of-Word + TFIDF and linear combination of Word2Vec vectors. The texts are fed to the input of the recurrent models were integer encoded and padded as sequences. Training of recurrent models was conducted using the embedding matrix built on the basis of pre-trained word embeddings.

\subsection{Synthetic Oversampling Algorithms}

The most direct method for solving the imbalance problem is random oversampling based on duplication of randomly selected vectors from the training sample.

In addition, we used the three popular algorithms of synthetic oversampling: SMOTE (Synthetic Minority Oversampling Technique) \cite{smote}, Borderline SMOTE \cite{borderline} and ADASYN \cite{adasyn}. All these algorithms consists of the following main steps \cite{Ah-Pine}:

\begin{enumerate}
    \item Randomly select the sample $s$.
    \item Determine $N_{s}$ as the nearest neighborhood of $s$.
    \item Select a random neighbor $s': s'\in N_{s}$.
    \item Create a new sample $s_{new}$:
\begin{equation}
    \it{s_{new}} = \it{s} + \alpha(\it{s'} - \it{s}), \alpha\in[0,1].
\end{equation}
    \item Repeat 1-4 until the desired number of samples is reached.
    \item Append synthetic examples to the training set.
\end{enumerate}

The difference between three considered algorithms is the way to select vectors for which the oversampling procedure is performed. The SMOTE algorithm selects a random example from the whole set of examples from current minority category. The Borderline SMOTE algorithm makes a choice only among the examples, most of the nearest neighbors of which do not belong to current minority category. In other words, the Borderline SMOTE works only with the samples lying on the boundary of the minority category. The ADASYN algorithm selects samples in accordance with a non uniform distribution. The probability of choosing a particular example directly depends on the number of points in the nearest neighborhood belonging to other categories.

To balance out the data received by recurrent networks we applied random oversampling of training samples of the minority classes. It is important to note that the SMOTE algorithm and its derivatives are hardly applicable to textual data. In the case of recurrent neural networks, unlike other applied machine learning methods, we performed oversampling of the original data, not the processed numerical vectors.

\section{Data Description}

We trained our models on the corpus of biographical texts available at \cite{corpus}. The process of the text corpus building were described in out previous works \cite{article}. The version of the corpus used in this study consists of 179 biographical texts from Russian Wikipedia \cite{Wiki} related to persons who live or lived in the XX-XXI centuries.

Sentences in the corpus have been manually tagged in accordance with the following taxonomy:

\begin{enumerate}
\item[1)] non-biographical facts;
\item[2)] unchangeable personal characteristics: birth, death, nationality, parenting information;
\item[3)] changeable personal characteristics: affiliation, education, family (marriage, children, etc.), occupation (position), personal events, professional events;
\item[4)] other biographical facts.
\end{enumerate}

We excluded the category ``nationality" from consideration, because it included only 3 examples. Also, the categories ``other biographical facts" and ``non-biographical facts" were excluded due to the fact that they do not contain information related to specific topics. Thus, we carried out the classification for the 10 topic categories. The corpus was randomly divided into training and test samples in the 80:20 ratio. The final distribution of sentences is presented in the Table 1.

\begin{table}[htbp]
\centering
\caption{The corpus of biographical facts}
\begin{tabular}{|c|c|c|}\hline
&{\textbf{Training sample}} & {\textbf{Test sample}} \\
\hline
\textbf{Total} & 1577 & 395\\
\hline
\textbf{Affiliation} & 171 & 29\\
\hline
\textbf{Birth} & 191 & 26\\
\hline
\textbf{Death} & 165 & 21\\
\hline
\textbf{Education} & 516 & 67\\
\hline
\textbf{Family} & 44 & 6\\
\hline
\textbf{Occupation} & 319 & 127\\
\hline
\textbf{Parenting} & 15 & 8\\
\hline
\textbf{Personal Events} & 23 & 15\\
\hline
\textbf{Professional Events} & 118 & 81\\
\hline
\textbf{Residence} & 15 & 15\\
\hline
\end{tabular}
\end{table}

A key observation in the dataset is the noticeable imbalance between the categories (Figure 1).

\begin{figure}
\centering
\includegraphics[scale=.4]{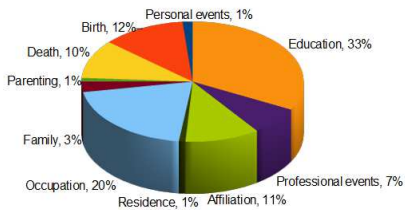}
\caption{Target distribution}
\end{figure}

\section{Experimental Results}

In this work, we used neural network models implemented using the Keras library \cite{Keras}. We chose the hyperbolic tangent activation for the hidden layers and the softmax for the output layer in recurrent networks. In feedforward networks, we used the sigmoid activation for hidden layers. The batch size was equal to 8. The optimization algorithm used is adaptive moment estimation (the Adam optimization). We carried out dropout regularization with a probability of 0.5. The number of neurons in the recurrent layers ranged from 16 up to 128 with a network depth of 1-2 hidden layers. SVM and KNN algorithms are implemented using the scikit-learn library \cite{scikit}. We used the count of nearest neighbors equal to 5 and the gamma value in the SVM algorithm equal to 0.001. Values of parameters are chosen experimentally. Oversampling methods are implemented using the imbalanced-learn library \cite{imblearn}. The source code of the models is available here \cite{code}.

Since the prediction results favor the majority categories at the expense of the minority classes we need to choose metrics that will be able to consider this feature. We use the F1 score with a macro average. It calculates as the unweighted mean of F1 score values for each category. Additionally, we specify the precision and recall scores which are calculated similarly for several categories.

In our work, we used the parameter $k$, expressed as a percentage, which specifies the number of generated synthetic examples. We calculate the number of synthetic examples $Size_n$ for the category $n$ as follows:

\begin{equation}
    \it{Size_n} = (\it{M} - \it{CurrentSize_n})*\it{k},
\end{equation}

where $M$ is the size of the largest category, $CurrentSize_n$ is the current size of the category $n$. We conducted our experiments with $k$ equal to 50, 75 and 100 percent.

The final results are shown in Figures 2-5. Full results are presented in Tables 2-4.

In most cases, oversampling algorithms improve the quality of text classification raising both the accuracy and the macro F1 score. The difference in the results is especially noticeable when we used the KNN and SVM methods. In most cases, the best results were achieved using the ADASYN algorithm and the random oversampling technique.

The least influence of oversampling is traced when using neural network methods. In some cases (for example, using the Bag-of-Words model and FNN), the results obtained after applying oversampling have worse values by F1 score than the results for the original dataset.

\begin{figure}[htbp]
\centering
\includegraphics[scale=.3]{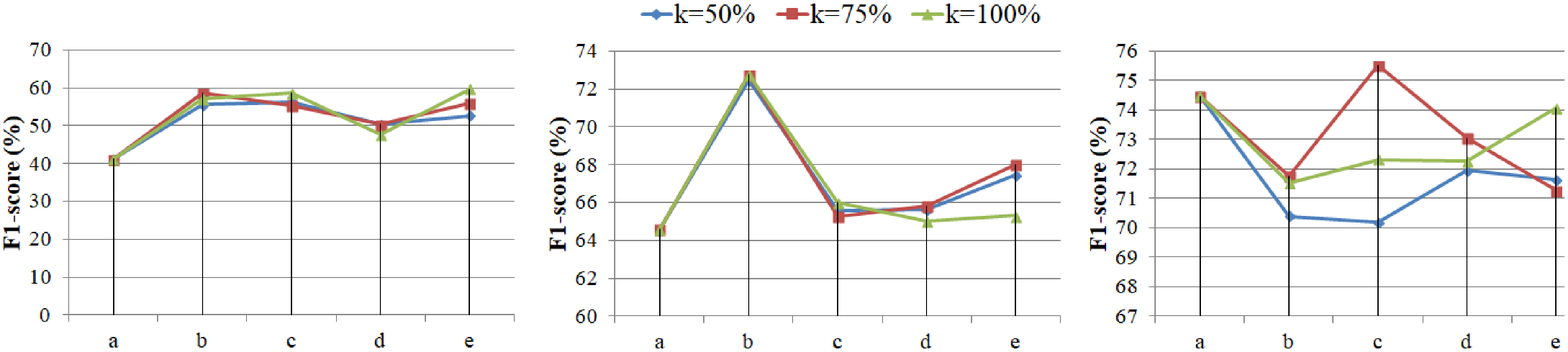}
\caption{Bag-of-Words. From left to right: KNN, SVM, FNN. Datasets: a - original dataset; b - random oversampling, c - SMOTE, d - borderline SMOTE, e - ADASYN.}
\end{figure}

\begin{figure}[htbp]
\centering
\includegraphics[scale=.3]{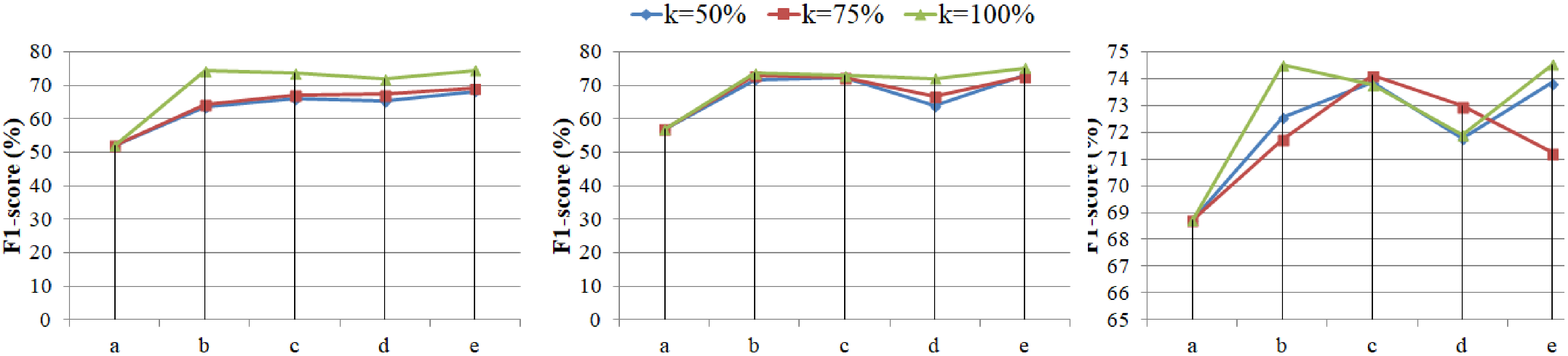}
\caption{Bag-of-Words + TFIDF. From left to right: KNN, SVM, FNN. Datasets: a - original dataset; b - random oversampling, c - SMOTE, d - borderline SMOTE, e - ADASYN.}
\end{figure}

\begin{figure}[htbp]
\centering
\includegraphics[scale=.3]{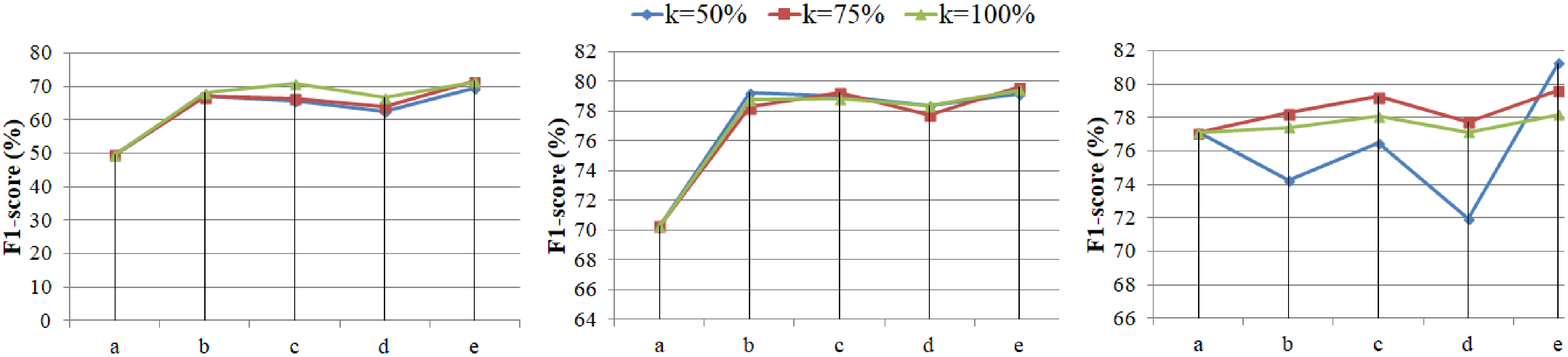}
\caption{Word2Vec (linear). From left to right: KNN, SVM, FNN. Datasets: a - original dataset; b - random oversampling, c - SMOTE, d - borderline SMOTE, e - ADASYN.}
\end{figure}

\begin{figure}[htbp]
\centering
\includegraphics[scale=0.5]{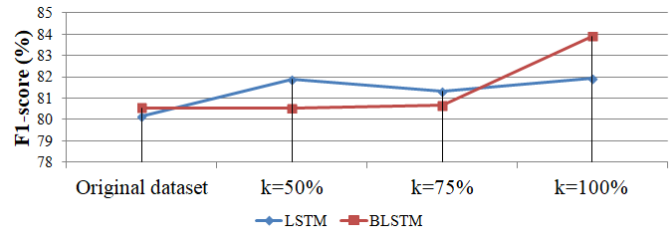}
\caption{Random oversampling results for recurrent models (Word2Vec (sequences)).}
\end{figure}

\begin{table}[htbp]
\caption{Experimental results ($k$=50\%)}
\begin{tabular}{|llcccccc|}
\hline
 & \multicolumn{1}{l|}{Metric(\%)} & \multicolumn{3}{l|}{Bag-of-Words} & \multicolumn{3}{l|}{Bag-of-Words + TFIDF} \\ \cline{3-8} 
 & \multicolumn{1}{l|}{} & \multicolumn{1}{l}{KNN} & \multicolumn{1}{l}{SVM} & \multicolumn{1}{l|}{FNN} & \multicolumn{1}{l}{KNN} & \multicolumn{1}{l}{SVM} & \multicolumn{1}{l|}{FNN} \\ \hline
Original training set & Accuracy & 50.63 & 74.94 & 79.24 & 60.51 & 66.84 & 75.95 \\
 & F1 score & 41.24 & 64.58 & 74.47 & 52.12 & 56.88 & 68.72 \\
 & Precision & 51.4 & 64.61 & 75.88 & 59.26 & 60.79 & 71.71 \\
 & Recall & 56.47 & 82.63 & 75.71 & 60.59 & 77.48 & 74.11 \\ \hline
Random over-sampling & Accuracy & 56.96 & 77.22 & 75.19 & 56.96 & 77.22 & 75.19 \\
 & F1 score & 55.52 & 72.46 & 70.4 & 55.52 & 72.46 & 70.4 \\
 & Precision & 63.13 & 70.72 & 71.34 & 63.13 & 70.72 & 71.34 \\
 & Recall & 70.69 & 88.02 & 77.36 & 70.69 & 88.02 & 77.36 \\ \hline
SMOTE & Accuracy & 61.01 & 73.67 & 76.96 & 61.01 & 73.67 & 76.96 \\
 & F1 score & 56.24 & 65.6 & 70.19 & 56.24 & 65.6 & 70.19 \\
 & Precision & 63.1 & 66.17 & 72.79 & 63.1 & 66.17 & 72.79 \\
 & Recall & 58.09 & 67.84 & 72.51 & 58.09 & 67.84 & 72.51 \\ \hline
Borderline SMOTE & Accuracy & 55.44 & 72.66 & 77.47 & 55.44 & 72.66 & 77.47 \\
 & F1 score & 50.41 & 65.63 & 71.95 & 50.41 & 65.63 & 71.95 \\
 & Precision & 56.88 & 65.31 & 72.63 & 56.88 & 65.31 & 72.63 \\
 & Recall & 61.97 & 82.17 & 77.13 & 61.97 & 82.17 & 77.13 \\ \hline
ADASYN & Accuracy & 58.23 & 74.43 & 74.43 & 58.23 & 74.43 & 74.43 \\
 & F1 score & 52.67 & 67.48 & 71.64 & 52.67 & 67.48 & 71.64 \\
 & Precision & 59.35 & 68.67 & 74.3 & 59.35 & 68.67 & 74.3 \\
 & Recall & 53.95 & 73.98 & 69.93 & 53.95 & 73.98 & 69.93 \\ \hline
 & \multicolumn{1}{l|}{Metric(\%)} & \multicolumn{3}{l|}{Word2Vec (linear)} & \multicolumn{3}{l|}{Word2Vec (sequences)} \\ \cline{3-8} 
 & \multicolumn{1}{l|}{} & \multicolumn{1}{l}{KNN} & \multicolumn{1}{l}{SVM} & \multicolumn{1}{l|}{FNN} & LSTM & \multicolumn{2}{c|}{BLSTM} \\ \hline
Original training set & Accuracy & 62.28 & 80.51 & 82.53 & 84.81 & \multicolumn{2}{c|}{84.05} \\
 & F1 score & 49.42 & 70.29 & 77.1 & 80.17 & \multicolumn{2}{c|}{80.55} \\
 & Precision & 58.02 & 68.83 & 79.14 & 80 & \multicolumn{2}{c|}{80.23} \\
 & Recall & 70.25 & 91.05 & 76.17 & 82.5 & \multicolumn{2}{c|}{81.57} \\ \hline
Random over-sampling & Accuracy & 71.39 & 83.54 & 80.11 & 82.28 & \multicolumn{2}{c|}{84.05} \\
 & F1 score & 67.23 & 79.24 & 74.26 & 81.87 & \multicolumn{2}{c|}{80.68} \\
 & Precision & 71.17 & 76.46 & 81.77 & 82.51 & \multicolumn{2}{c|}{82.33} \\
 & Recall & 69 & 90.34 & 80 & 81.57 & \multicolumn{2}{c|}{81.37} \\ \hline
SMOTE & Accuracy & 70.38 & 82.53 & 81.21 & - & \multicolumn{2}{c|}{-} \\
 & F1 score & 65.55 & 78.98 & 76.48 & - & \multicolumn{2}{c|}{-} \\
 & Precision & 74.74 & 75.86 & 77.46 & - & \multicolumn{2}{c|}{-} \\
 & Recall & 62.48 & 90.17 & 79.35 & - & \multicolumn{2}{c|}{-} \\ \hline
Borderline SMOTE & Accuracy & 67.34 & 82.28 & 76.46 & - & \multicolumn{2}{c|}{-} \\
 & F1 score & 62.7 & 78.38 & 71.95 & - & \multicolumn{2}{c|}{-} \\
 & Precision & 70.28 & 75.12 & 78.39 & - & \multicolumn{2}{c|}{-} \\
 & Recall & 65.43 & 90.96 & 72.88 & - & \multicolumn{2}{c|}{-} \\ \hline
ADASYN & Accuracy & 72.91 & 82.78 & 80.25 & - & \multicolumn{2}{c|}{-} \\
 & F1 score & 69.34 & 79.19 & 81.24 & - & \multicolumn{2}{c|}{-} \\
 & Precision & 76.87 & 75.99 & 82.62 & - & \multicolumn{2}{c|}{-} \\
 & Recall & 66.28 & 90.94 & 82.91 & - & \multicolumn{2}{c|}{-} \\ \hline
\end{tabular}
\end{table}

\begin{table}[htbp]
\caption{Experimental results ($k$=75\%)}
\begin{tabular}{|llcccccc|}
\hline
 & \multicolumn{1}{l|}{Metric(\%)} & \multicolumn{3}{l|}{Bag-of-Words} & \multicolumn{3}{l|}{Bag-of-Words + TFIDF} \\ \cline{3-8} 
 & \multicolumn{1}{l|}{} & \multicolumn{1}{l}{KNN} & \multicolumn{1}{l}{SVM} & \multicolumn{1}{l|}{FNN} & \multicolumn{1}{l}{KNN} & \multicolumn{1}{l}{SVM} & \multicolumn{1}{l|}{FNN} \\ \hline
Random over-sampling & Accuracy & 60.76 & 77.72 & 75.95 & 67.34 & 77.72 & 75.95 \\
 & F1 score & 58.64 & 72.75 & 71.77 & 64.41 & 72.88 & 71.37 \\
 & Precision & 64.32 & 70.96 & 70.95 & 69.69 & 70.98 & 71.95 \\
 & Recall & 77.93 & 88.01 & 81.66 & 64.32 & 84.64 & 79.35  \\ \hline
SMOTE & Accuracy & 64.05 & 73.16 & 78.73 & 69.11 & 77.47 & 76.96 \\
 & F1 score & 55.32 & 65.29 & 75.51 & 67.19 & 72.34 & 74.11 \\
 & Precision & 61.8 & 65.93 & 77.93 & 76.49 & 70.24 & 73.55 \\
 & Recall & 55.47 & 67.54 & 75.74 & 62.59 & 85.16 & 80.34 \\ \hline
Borderline SMOTE & Accuracy & 61.01 & 73.16 & 77.97 & 70.38 & 73.42 & 78.48 \\
 & F1 score & 50.33 & 65.83 & 73.06 & 67.28 & 66.64 & 72.97 \\
 & Precision & 57.5 & 65.47 & 74.45 & 74.57 & 62.46 & 73.94 \\
 & Recall & 62.32 & 82.18 & 75 & 63.38 & 88.61 & 77.89  \\ \hline
ADASYN & Accuracy & 63.8 & 73.92 & 75.7 & 69.87 & 77.22 & 78.99 \\
 & F1 score & 55.94 & 68.03 & 71.27 & 69.04 & 72.81 & 71.23 \\
 & Precision & 61.01 & 68.42 & 75.01 & 75.53 & 71.02 & 72.12 \\
 & Recall & 63.4 & 69.08 & 69.27 & 65.58 & 84.82 & 77.96 \\ \hline
 & \multicolumn{1}{l|}{Metric(\%)} & \multicolumn{3}{l|}{Word2Vec (linear)} & \multicolumn{3}{l|}{Word2Vec (sequences)} \\ \cline{3-8} 
 & \multicolumn{1}{l|}{} & \multicolumn{1}{l}{KNN} & \multicolumn{1}{l}{SVM} & \multicolumn{1}{l|}{FNN} & LSTM & \multicolumn{2}{c|}{BLSTM} \\ \hline
Random oversampling & Accuracy & 71.39 & 82.53 & 77.22 & 85.82 & \multicolumn{2}{c|}{84.05} \\
 & F1 score & 66.93 & 78.28 & 75.7 & 81.31 & \multicolumn{2}{c|}{80.53} \\
 & Precision & 71.92 & 75.64 & 78.81 & 78.92 & \multicolumn{2}{c|}{80.83} \\
 & Recall & 68.01 & 89.55 & 74.26 & 86.21 & \multicolumn{2}{c|}{80.74} \\ \hline
SMOTE & Accuracy & 71.65 & 83.04 & 77.97 & - & \multicolumn{2}{c|}{-} \\
 & F1 score & 66.37 & 79.25 & 74.98 & - & \multicolumn{2}{c|}{-} \\
 & Precision & 73.82 & 76.37 & 78.13 & - & \multicolumn{2}{c|}{-} \\
 & Recall & 63.9 & 90.19 & 76.7 & - & \multicolumn{2}{c|}{-} \\ \hline
Borderline SMOTE & Accuracy & 68.35 & 82.03 & 79.24 & - & \multicolumn{2}{c|}{-} \\
 & F1 score & 63.99 & 77.75 & 72.2 & - & \multicolumn{2}{c|}{-} \\
 & Precision & 70.8 & 74.45 & 71.66 & - & \multicolumn{2}{c|}{-} \\
 & Recall & 67.19 & 90.73 & 82.05 & - & \multicolumn{2}{c|}{-} \\ \hline
ADASYN & Accuracy & 75.19 & 83.8 & 83.88 & - & \multicolumn{2}{c|}{-} \\
 & F1 score & 71.53 & 79.62 & 79.62 & - & \multicolumn{2}{c|}{-} \\
 & Precision & 77.62 & 76.43 & 76.43 & - & \multicolumn{2}{c|}{-} \\
 & Recall & 69.82 & 91.32 & 91.32 & - & \multicolumn{2}{c|}{-} \\ \hline
\end{tabular}
\end{table}

\begin{table}[htbp]
\caption{Experimental results ($k$=100\%)}
\begin{tabular}{|llcccccc|}
\hline
 & \multicolumn{1}{l|}{Metric(\%)} & \multicolumn{3}{l|}{Bag-of-Words} & \multicolumn{3}{l|}{Bag-of-Words + TFIDF} \\ \cline{3-8} 
 & \multicolumn{1}{l|}{} & \multicolumn{1}{l}{KNN} & \multicolumn{1}{l}{SVM} & \multicolumn{1}{l|}{FNN} & \multicolumn{1}{l}{KNN} & \multicolumn{1}{l}{SVM} & \multicolumn{1}{l|}{FNN} \\ \hline
Random oversampling & Accuracy & 58.73 & 77.72 & 74.68 & 69.62 & 79.24 & 80 \\
 & F1 score & 57.11 & 72.75 & 71.52 & 66.37 & 73.62 & 74.52 \\
 & Precision & 63.77 & 70.96 & 70.83 & 71.01 & 71.84 & 72.76 \\
 & Recall & 71.91 & 88.01 & 77.02 & 67.09 & 85.23 & 82.15 \\ \hline
SMOTE & Accuracy & 62.78 & 73.92 & 78.23 & 67.34 & 78.23 & 78.73 \\
 & F1 score & 58.61 & 65.99 & 72.32 & 66 & 72.91 & 73.78 \\
 & Precision & 66.27 & 66.3 & 73.35 & 73.94 & 70.77 & 72.57 \\
 & Recall & 58.54 & 73.01 & 79.7 & 61.65 & 85.23 & 84.4 \\ \hline
Borderline SMOTE & Accuracy & 58.22 & 71.39 & 76.71 & 70.38 & 73.67 & 76.46 \\
 & F1 score & 47.78 & 65.01 & 72.26 & 67.82 & 72.08 & 71.88 \\
 & Precision & 55.18 & 64.88 & 73.36 & 72.64 & 82.89 & 71.84 \\
 & Recall & 57.45 & 74.7 & 76.94 & 66 & 67.21 & 77.18 \\ \hline
ADASYN & Accuracy & 63.8 & 72.41 & 78.73 & 71.65 & 79.24 & 78.73 \\
 & F1 score & 59.73 & 65.3 & 74.07 & 70 & 75.19 & 74.53 \\
 & Precision & 63.74 & 65.68 & 75.32 & 77.3 & 72.15 & 74.98 \\
 & Recall & 60.07 & 72.21 & 74.18 & 66.07 & 87.8 & 80.23 \\ \hline
 & \multicolumn{1}{l|}{Metric(\%)} & \multicolumn{3}{l|}{Word2Vec (linear)} & \multicolumn{3}{l|}{Word2Vec (sequences)} \\ \cline{3-8} 
 & \multicolumn{1}{l|}{} & \multicolumn{1}{l}{KNN} & \multicolumn{1}{l}{SVM} & \multicolumn{1}{l|}{FNN} & LSTM & \multicolumn{2}{c|}{BLSTM} \\ \hline
Random oversampling & Accuracy & 71.65 & 83.29 & 81.27 & 85.82 & \multicolumn{2}{c|}{85.06} \\
 & F1 score & 68.16 & 78.77 & 77.42 & 81.91 & \multicolumn{2}{c|}{83.91} \\
 & Precision & 72.13 & 76.19 & 80.54 & 82.26 & \multicolumn{2}{c|}{83.95} \\
 & Recall & 69.33 & 89.97 & 78.53 & 82.72 & \multicolumn{2}{c|}{85.19} \\ \hline
SMOTE & Accuracy & 72.91 & 82.78 & 81.01 & - & \multicolumn{2}{c|}{-} \\
 & F1 score & 70.79 & 78.85 & 78.07 & - & \multicolumn{2}{c|}{-} \\
 & Precision & 76.34 & 75.68 & 81.06 & - & \multicolumn{2}{c|}{-} \\
 & Recall & 68.51 & 90.08 & 78.81 & - & \multicolumn{2}{c|}{-} \\ \hline
Borderline SMOTE & Accuracy & 70.89 & 82.27 & 82.03 & - & \multicolumn{2}{c|}{-} \\
 & F1 score & 66.71 & 78.38 & 77.14 & - & \multicolumn{2}{c|}{-} \\
 & Precision & 71.45 & 75.12 & 78.26 & - & \multicolumn{2}{c|}{-} \\
 & Recall & 71.62 & 90.96 & 80.82 & - & \multicolumn{2}{c|}{-} \\ \hline
ADASYN & Accuracy & 73.67 & 83.29 & 82.28 & - & \multicolumn{2}{c|}{-} \\
 & F1 score & 71.4 & 79.42 & 78.16 & - & \multicolumn{2}{c|}{-} \\
 & Precision & 77.22 & 76.23 & 78.45 & - & \multicolumn{2}{c|}{-} \\
 & Recall & 69.76 & 91.12 & 79.17 & - & \multicolumn{2}{c|}{-} \\ \hline
\end{tabular}
\end{table}

\section{Conclusion}

While solving text classification tasks, the class imbalance problem often arises. This problem can lead to degradation in the classification performance, but oversampling methods can partially handle this imbalance.

In our future work, we plan to conduct a study of oversampling methods for different NLP tasks. In addition, we plan to compare the results obtained for the multi-class topic classification for this corpus with the results for other corpora.

\end{document}